\pdfoutput=1
\documentclass[11pt]{article}

\usepackage[final]{acl}

\usepackage{times}
\usepackage{latexsym}

\usepackage[T1]{fontenc}
\usepackage[utf8]{inputenc}

\usepackage{microtype}

\usepackage{inconsolata}

\usepackage{algorithm}
\usepackage{algorithmic}
\usepackage{amsmath}
\usepackage{booktabs}
\usepackage{multirow}
\usepackage{subcaption}
\usepackage{enumitem}
\usepackage{mathtools}
\usepackage{makecell}
\usepackage{hyperref}
\usepackage{xcolor}
\usepackage[normalem]{ulem}
\usepackage{tablefootnote}
\usepackage{float}

\usepackage{color-edits}
\usepackage{comment}
\usepackage{soul}

\usepackage{amssymb}% http://ctan.org/pkg/amssymb
\usepackage{pifont}% http://ctan.org/pkg/pifont
\newcommand{\cmark}{\ding{51}}%
\newcommand{\xmark}{\ding{55}}%

\title{\textsc{Reviewer2}: Optimizing Review Generation Through Prompt Generation}

\author{Zhaolin Gao, Kianté Brantley, and Thorsten Joachims \\
Department of Computer Science, Cornell University \\
Ithaca, NY, USA \\
\texttt{\{zg292, kdb82\}@cornell.edu}, \{tj\}@cs.cornell.edu}

\begin{document}
\maketitle
\begin{abstract}
Recent developments in LLMs offer new opportunities for assisting authors in improving their work. In this paper, we envision a use case where authors can receive LLM-generated reviews that uncover weak points in the current draft. While initial methods for automated review generation already exist, these methods tend to produce reviews that lack detail, and they do not cover the range of opinions that human reviewers produce. To address this shortcoming, we propose an efficient two-stage review generation framework called \textsc{Reviewer2}. Unlike prior work, this approach explicitly models the distribution of possible aspects that the review may address. We show that this leads to more detailed reviews that better cover the range of aspects that human reviewers identify in the draft. As part of the research, we generate a large-scale review dataset of 27k papers and 99k reviews that we annotate with aspect prompts, which we make available as a resource for future research.
\end{abstract}

\section{Introduction}

Asking fellow group members to critique a draft is widely regarded as a valuable way of improving scientific writing, and the lack of access to such peers outside of well-resourced research groups is a key source of inequality \cite{matthew-effect, global_citation, inequalities_in_science}.
Furthermore, even in well-resourced groups, the frequency with which authors can receive feedback is limited. In this paper, we thus develop techniques for generating automated feedback via LLMs to aid authors in enhancing the quality of their work before it enters the formal peer review.
This helps level the playing field, and it promises to reduce pressure on the peer review process \cite{lee2012} after experiencing exponential increases in submissions \cite{bjork2013solomon, bornmann2014growth, kelly2014sadeghieh}.

The ability of LLMs to reason about complex tasks gives them the potential to provide automated feedback on papers~\cite{liu2023reviewergpt, liang2023large}.
A key asset is that we already have substantial amounts of supervised data from peer reviews \cite{kang-etal-2018-dataset, yuan2021automate, shen-etal-2022-mred, dycke-etal-2023-nlpeer}, containing paper-review pairs across different years, venues, and subjects. 
Prior approaches to review generation \cite{yuan2021automate, Lin_2023} focus on fine-tuning a pre-trained language model based on these datasets. However, unlike typical instruction following tasks \cite{ouyang2022training, touvron2023llama}, we argue that open-ended review generation is under-specified in a way that makes it difficult to align language models for instruction following. In particular, asking an LLM to generate a review without specifying which aspects of the paper to focus on exposes the model to substantial uncertainty. This leads to shortcomings along the following dimensions:

\begin{table*}[t]\centering
\vskip -0.3cm
\resizebox{0.93\linewidth}{!}{
\begin{tabular}{p{0.33\linewidth}p{0.33\linewidth}p{0.33\linewidth}} 
\midrule[0.15ex]
  \multicolumn{1}{c}{\textbf{Training Step 500}} & \multicolumn{1}{c}{\textbf{Training Step 1000}} & \multicolumn{1}{c}{\textbf{Training Step 2000}}\\ 
  \hline
   The paper proposes a simple and efficient differentiable data generation pipeline.
   &
   The authors have done extensive experiments to validate the effectiveness of the proposed method.
   &
   The paper is well-written and straightforward. The method is technically sound.
   \\
\midrule[0.15ex]
\end{tabular}}
\vskip -0.4cm
\caption{Generated reviews from different steps using a generic prompt ($1$ epoch $\approx$ $1000$ steps).}
\label{tab:specificity}
\vskip -0.2cm
\end{table*}

\textbf{Specificity.}
Peer reviews exhibit varying levels of specificity from general (e.g., "the paper is technically sound.") to precise (e.g., "the paper has a good theoretical basis based on the derivation in section 3."). A good review should provide detailed justifications for its assessment, especially when stating the weaknesses of the paper \cite{yuan2021automate}. In addition, justifications make the review more constructive as they provide direct instructions on how to improve the paper~\cite{xiong-litman-2011-automatically}. However, our experiments reveal that standard fine-tuning diminishes the specificity of the generated reviews. An example is shown in Table~\ref{tab:specificity} where we generate reviews based on a model that is fine-tuned over increasing numbers of training steps. The generated review is significantly more generic at step $2000$ compared to the one at step $500$.

\textbf{Coverage and Control.}
Different human reviewers are likely to focus on different aspects of a paper. An automated review-generation system should thus cover the range of issues that human reviewers may identify. We find that standard fine-tuning of LLMs for review generation often leads to a form of regression-to-the-mean, where the generated reviews do not cover the full range of aspects. We argue that an ideal system should actively control coverage, and give authors the ability to ask for feedback on specific aspects. 
%\kbcomment{(Add a contrasting statement with generic prompts), for example, "Existing datasets use generic prompts that do not cover aspects which hinder the development of a controllable review generation system that can review different aspects of a paper draft.}
%The absence of aspect-based prompts in existing datasets hinders the development of a controllable review generation system.

To address these issues, we propose an efficient two-stage review generation framework for papers called \textsc{Reviewer2}. \textsc{Reviewer2} includes two fine-tuned language models. The first LLM analyzes the paper and produces a set of aspects that the reviews should focus on. Each of these aspects takes the form of a prompt that is the input for the second stage. The second LLM generates a review based on the paper and the aspect prompt. We implement \textsc{Reviewer2} based on LongLoRA~\cite{chen2023longlora} to enable 32k context length, avoiding the use of extractive summaries of the paper that was necessary in prior work due to limitations in context length \cite{gehrmann2018bottomup, chen-bansal-2018-fast, dou-etal-2021-gsum, yuan2021automate, Lin_2023}. 
%Using extracted summaries removes details about the paper, limiting the quality of the generated review.

Unfortunately, existing peer-review datasets do not include aspect prompts, providing insufficient data for training either stage of \textsc{Reviewer2}. To address this issue, we develop a Prompt Generation with Evaluation (\textit{PGE}) pipeline to generate a variety of high-quality aspect prompts. \textit{PGE} generates prompts given the review and uses a self-evaluation step to ensure the quality of the generated prompts. Based on \textit{PGE}, we construct a large-scale review dataset of 27k papers and 99k reviews from six different venues with corresponding aspect prompts. 

Extensive experiments on multiple venues demonstrate that our \textsc{Reviewer2} framework trained on \textit{PGE}-generated aspect prompts substantially outperforms existing methods in terms of review quality, specificity, and coverage. The major contributions of this paper are summarized below:
\begin{itemize}[leftmargin=*,noitemsep,topsep=0pt]
\item We propose \textsc{Reviewer2}, a novel framework for joint aspect prompt and review generation that improves coverage and enables control.
\item We implement \textsc{Reviewer2} based on LongLoRA, enabling 32k context length with low memory requirement for fine-tuning.
\item We design two new metrics for evaluating specificity and coverability. We compare \textsc{Reviewer2} with various baseline methods and find that it substantially improves review generation.
\item We propose \textit{PGE}, a novel pipeline for augmenting existing review datasets with aspect prompts, and we construct the first large-scale peer review dataset that includes aspect prompts.
\end{itemize}

\section{Related Work}

\textbf{Instruction generation and tuning.}
Previous works have demonstrated the efficacy of instruction fine-tuning in enhancing both task performance and adaptability to unseen tasks \cite{wei2022finetuned, sanh2022multitask, ouyang2022training}. However, these approaches depend heavily on human-written instruction data, which is often constrained in terms of quantity and diversity. Several works have explored using large language models (LLMs) to automatically generate instructions. \citet{honovich2022unnatural} prompts a language model with seed examples of instructions to generate additional instructions, inputs, and outputs. \citet{wang2023selfinstruct} adopts a similar approach while filtering the generated instructions to ensure diversity and quality.

\textbf{Self-alignment.}
Self-alignment of LLMs is an emerging area of research that utilizes the model to improve itself and align with human values with minimal human supervision. This field primarily consists of two approaches: unsupervised data generation and post-hoc output refinement. In \citet{li2023selfalignment}, prompts and responses are generated according to a small set of human-written principles, while \citet{sun2023principledriven} focuses on generating synthetic prompts derived from human-written documents. On the other hand, \citet{madaan2023selfrefine} employs an iterative process to refine its output through generated feedback.

\textbf{Automation in peer review.}
Automated systems have played a significant role in various aspects of the review process. Numerous algorithms \cite{stelmakh2019peerreview4all, kobren2019paper, cohan2020specter} have been developed to evaluate the expertise of potential reviewers, optimizing reviewer-paper assignments. In addition, several algorithms have been proposed to ensure the submissions adhere to appropriate guidelines, such as plagiarism detection \cite{folt2020plagiarism} and desk rejection prediction \cite{8791152}. Recently, efforts have been directed towards the development of algorithms for review generation \cite{yuan2021automate, Lin_2023}, leveraging papers as input and fine-tuning on LLMs for review generation.

\begin{figure}[t]
\centering
\includegraphics[scale=0.6, trim={290 10 290 15},clip]{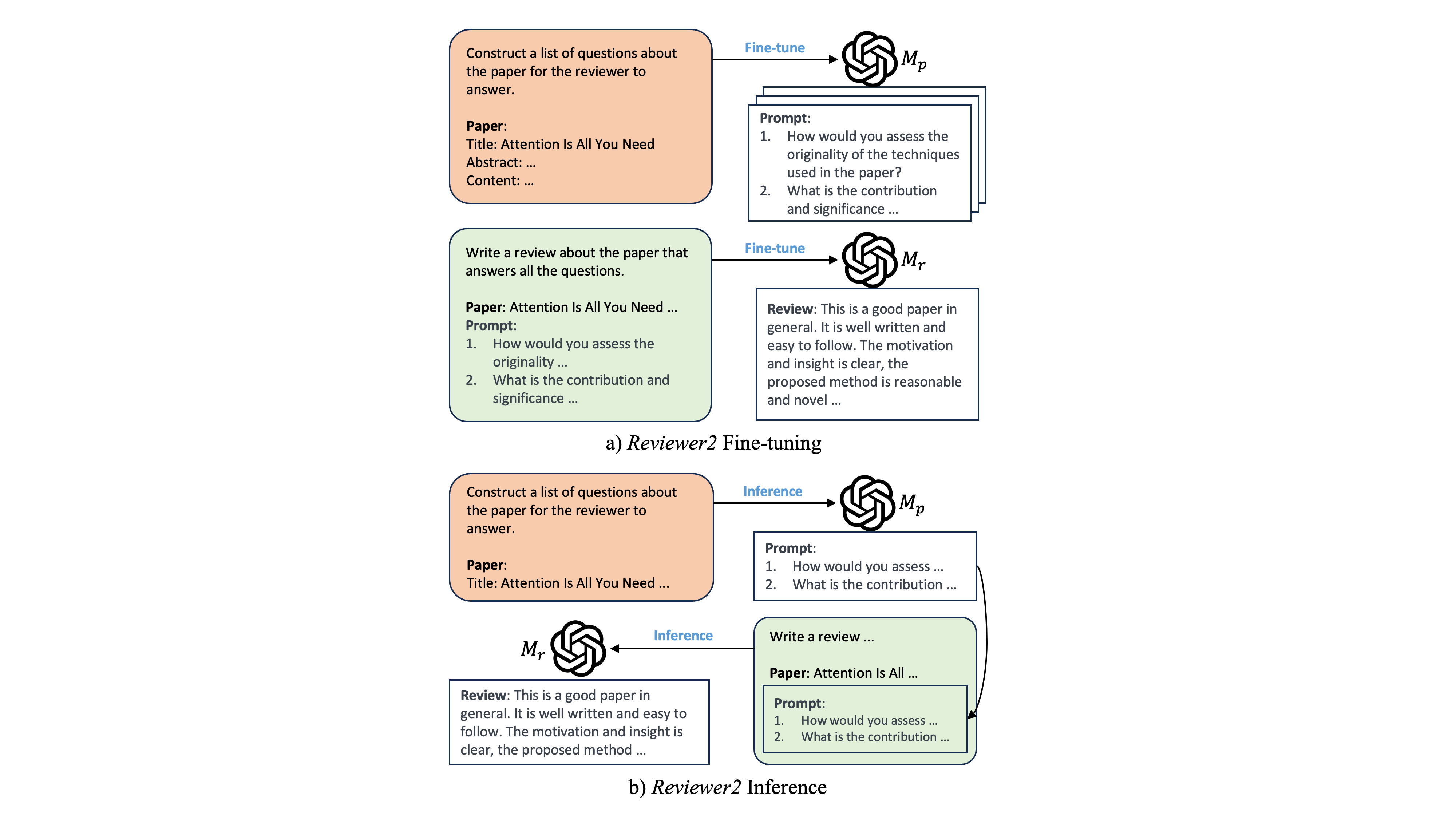}
\vskip -0.3cm
\caption{Illustrations of \textsc{Reviewer2}. \textbf{a)} \textsc{Reviewer2} fine-tunes two models: $M_p$ generates aspect prompts based on paper, and $M_r$ generates reviews based on the paper and a prompt. \textbf{b)} \textsc{Reviewer2} utilizes a two-stage inference to generate an aspect prompt and generate the review based on the generated prompt.}
\label{fig:reviewer2}
\vskip -0.3cm
\end{figure}

\begin{figure*}[t]
\centering
\includegraphics[scale=0.65, trim={130 180 140 180},clip]{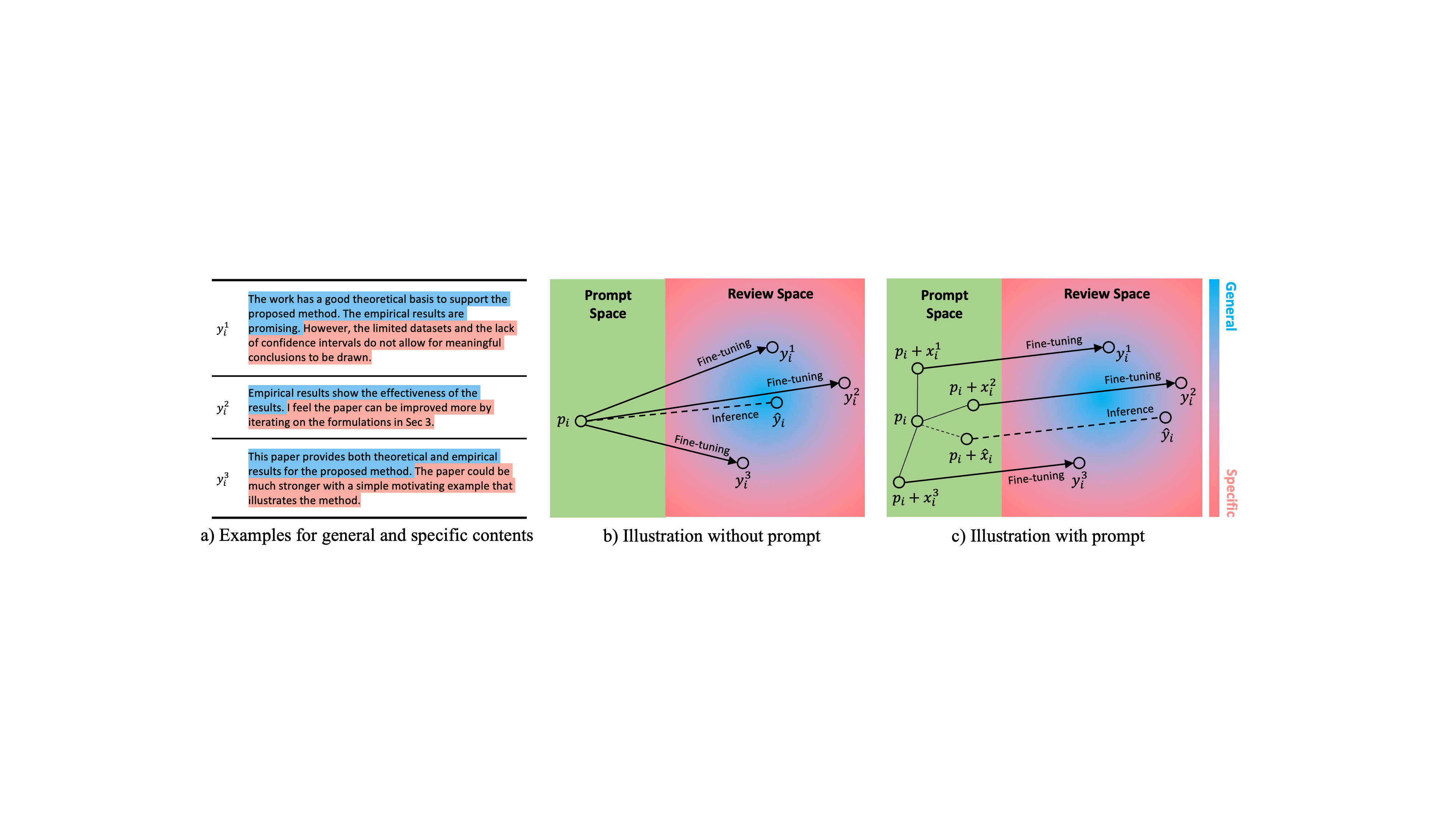}
\vskip -0.2cm
\caption{Illustrations of the effect of aspect prompts. \textbf{a)} General content is highlighted in blue, while specific content is highlighted in red. \textbf{b)} Fine-tuning without aspect prompts causes the generated contents to be general during inference. \textbf{c)} Fine-tuning with aspect prompts allows specific content generation during inference.}
\label{fig:prompt_illustration}
\vskip -0.2cm
\end{figure*}

\begin{figure}[t]
\centering
\includegraphics[scale=0.59, trim={283 80 290 80},clip]{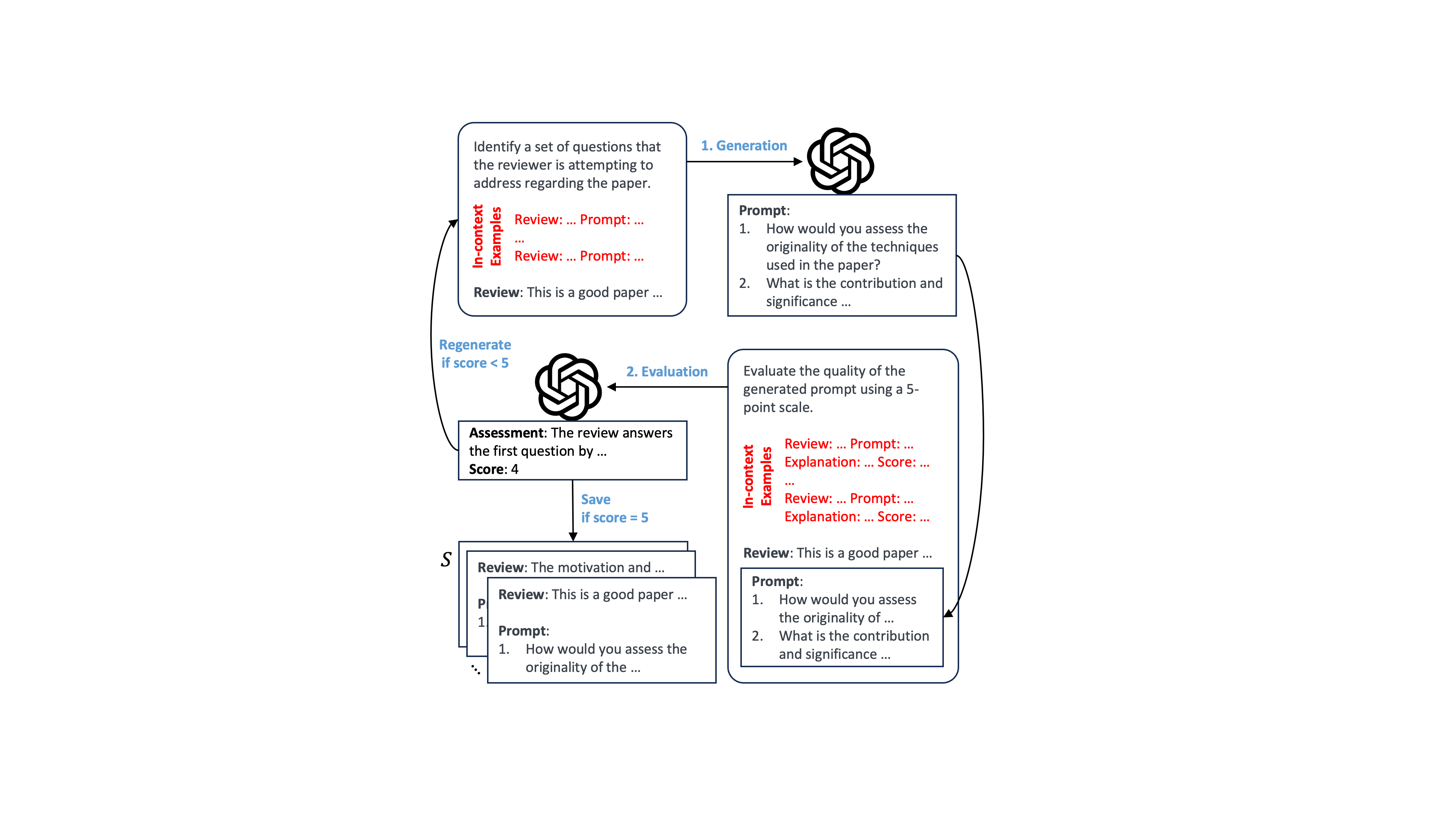}
\vskip -0.3cm
\caption{\textit{PGE} includes two steps: generation and evaluation. The prompt is regenerated if the score is below $5$ on a $5$-point scale, otherwise, it is saved to $S$.}
\label{fig:pge}
\vskip -0.3cm
\end{figure}

\section{\textsc{Reviewer2} for Review Generation}  \label{reviewer2}

In this section, we introduce our \textsc{Reviewer2} pipeline for generating reviews. The key idea is to insert explicit control into the pipeline to ensure that the generated reviews cover the full range of aspects that human reviewers may comment on. We demonstrate that this improves both coverage and specificity of the generated reviews.

Figure~\ref{fig:reviewer2}(a) illustrates how we train the two stages of \textsc{Reviewer2}. For the first stage, we fine-tune an LLM $$M_p: p \rightarrow \{x^1, ..., x^k\}$$ to produce a set of aspect prompts $x^1,...x^k$ for paper $p$ that cover the aspects that a reviewer may comment on for this paper. For the second stage of \textsc{Reviewer2}, we fine-tune another LLM $$M_r: (p, x) \rightarrow y$$ to produce a review $y$ for paper $p$ that addresses aspect $x$. When generating a review for a new paper $p'$, we first query $M_p$ for an aspect prompt $x$. We then query $M_r$ to produce a review $y$ for the generated aspect prompt. This inference process is depicted in Figure~\ref{fig:reviewer2}(b).
We will provide evidence that this two-stage pipeline not only provides explicit control of aspect coverage, it also avoids a type of regression-to-the-mean~\cite{rtm} that makes single-stage pipelines produce generic reviews with little specificity. 

An illustrative example is shown in Figure~\ref{fig:prompt_illustration} which contains three reviews, $\{y_i^1, y_i^2, y_i^3\}$, for paper $p_i$. All three reviews comment on either or both theoretical and empirical justifications, representing the general aspects. However, the reviews provide different suggestions for improvement, which are considered as specific parts. We find that a single-stage pipeline that is trained without aspect prompts tends to only generate the general components of the review, as illustrated in Figure~\ref{fig:prompt_illustration}(b), since such "mean reviews" align closely with all three reviews. On the other hand, by adding aspect prompts $\{x_i^1, x_i^2, x_i^3\}$ derived from the paper, the augmentation diversifies the aspects that are addressed, aligning it more effectively with the variability seen in the human reviews. Note that the prompt space now better captures the variability between reviewers, which reduces the noise when mapping to generated reviews. % when conditioned on the aspect prompt.
%Note the prompt space now captures the variability between reviewers, making the mapping to generated reviews less noisy once it is conditioned on the aspect prompt. 
This reduction in noise enables the generation of more specific reviews, $\hat{y}_i$, during inference as shown in Figure~\ref{fig:prompt_illustration}(c). The geometric intuition behind our illustration is detailed in Appendix~\ref{app:illustration}.

To enable efficient long context fine-tuning and inference, we adapt $\mathrm{LoRA}^+$ and $\mathrm{S}^2$-Attn from \citet{chen2023longlora}. $\mathrm{LoRA}^+$ extends on top of LoRA~\cite{hu2021lora} by making the embedding and normalization layers trainable, and $\mathrm{S}^2$-Attn groups input tokens to address the quadratic complexity of self-attention.

\section{Review Dataset with Aspect Prompts}

Training \textsc{Reviewer2} requires a dataset of papers and reviews that is augmented with aspect prompts. 
While there is ample data on papers and their associated reviews, these datasets contain generic review prompts that do not capture which aspects the human reviewer chose to focus on. 
We therefore developed the following methodology for augmenting existing review datasets with aspect prompts.

The result is the first review dataset that is annotated with aspect prompts, and we make this dataset available as a new resource. It consists of up-to-date crawls of publicly available reviews from {NeurIPS}
%\footnote{Conference on Neural Information Processing Systems} 
and {ICLR}
%\footnote{International Conference on Learning Representations}
, and we also augment the datasets from PeerRead~\cite{kang2018dataset} and NLPeer~\cite{dycke-etal-2023-nlpeer}.

\begin{table*}[t]\centering
\caption{Dataset Statistics \label{tab:dataset}}
\vskip -0.3cm
\resizebox{\linewidth}{!}{
\begin{tabular}{rccccccc} 
\midrule[0.15ex]
            & \textbf{CONLL-16} & \textbf{ACL-17}  & \textbf{COLING-20}  & \textbf{ARR-22} & \textbf{ICLR-17-23} & \textbf{NeurIPS-16-22} & \textbf{total}
\\  \hline
\# papers    & 22                & 137              & 89                  & 476             & 16,327              & 10,754                 & 27,805
\\
\# words per paper & 4,325              & 4,679           & 4,230                & 4,850           & 6,959                & 5,236                   & 6,229
\\
\# reviews   & 39                & 275              & 112                 & 684             & 58,933              & 39,684                 & 99,727
\\
\# words per review & 418              & 440           & 414                & 397           & 512                & 482                   & 487
\\
\# prompts   & 37                & 270              & 108                 & 676             & 58,107              & 38,762                 & 97,960
\\
\# words per prompt & 56              & 60           & 45                & 46           & 52                & 51                   & 53
\\
\% accepted & 50\%              & 67\%             & 93\%                & 100\%           & 32\%                & 98\%                   & 55\%
\\
domain & NLP/CL              & NLP/CL           & NLP/CL                & NLP/CL           & ML                & ML                   & multi
\\  \midrule[0.15ex]
\end{tabular}}
\vskip -0.5cm
\end{table*}

\begin{table}[t]\centering
\caption{Dataset Comparison \label{tab:dataset_comp}}
\vskip -0.3cm
\resizebox{\linewidth}{!}{
\begin{tabular}{cccc} 
\midrule[0.15ex]
                  & \# papers & \# reviews  & prompts
\\  \hline
PeerRead & \multirow{2}{*}{3,$006^*$} & \multirow{2}{*}{10,770} & \multirow{2}{*}{\xmark}
\\
\cite{kang2018dataset} &  &  &
\\
ASAP-Review & \multirow{2}{*}{8,877} & \multirow{2}{*}{28,119} & \multirow{2}{*}{\xmark}
\\
\cite{yuan2021automate} &  &  &
\\
MReD & \multirow{2}{*}{7,894} & \multirow{2}{*}{30,764} & \multirow{2}{*}{\xmark}
\\
\cite{shen-etal-2022-mred} &  &  &
\\
NLPeer & \multirow{2}{*}{5,672} & \multirow{2}{*}{11,515} & \multirow{2}{*}{\xmark}
\\
\cite{dycke-etal-2023-nlpeer} &  &  &
\\
Ours & 27,805 & 99,727 & \cmark
\\  \midrule[0.15ex]
\end{tabular}}
\raggedright{\footnotesize{*Number of papers that have reviews.}}
\vskip -0.5cm
\end{table}

\subsection{PGE: Prompt Generation with Evaluation} \label{prompt_gen}
In order to generate the corresponding prompt for each review, we propose Prompt Generation with Evaluation (\textit{PGE}) pipeline consisting of a generation step and an evaluation step, as shown in Figure~\ref{fig:pge}. Specifically, given a set of $m$ papers $P=\{p_1, p_2, ..., p_m\}$ and corresponding reference reviews $Y=\{y_i^n | 1 \leq i \leq m, 1 \leq n \leq n_i\}$ where $n_i$ is the number of reviews for paper $i$, the goal of the pipeline is to generate a set of prompts $X=\{x_i^n | 1 \leq i \leq m, 1 \leq n \leq n_i\}$ that one prompt corresponds to one review.

For a review $y_i^n$, the generation step generates a prompt, $x_i^n$, and the evaluation step evaluates the generated prompt based on a 5-point scale. If $x_i^n$ achieves a score of $5$, the pair $(x_i^n, y_i^n)$ is stored in the set $S$, $S=S\cup\{(x_i^n, y_i^n)\}$, otherwise the prompt is regenerated. This two-step iterative approach resolves the problem of the absence of ground-truth prompts for reviews and ensures the quality of prompt generation without human supervision. The prompts we used for generation and evaluation are shown in Appendix \ref{app:prompts}.

\textbf{Prompt Generation.}
We initialize $S$ with human-annotated examples that will be used as initial in-context examples during generation. To construct these examples, we use Llama-2-70B-Chat~\cite{touvron2023llama} to generate prompts for a randomly selected subset of $100$ reviews in a zero-shot fashion. Then, we manually refine the prompts by removing irrelevant questions, adding missing questions that are covered in the review, and refining to align with the open-ended format of review questions. An example of a review-prompt pair is shown in Appendix \ref{app:exp_prompt}.

To enhance the performance of prompt generation, we apply in-context learning (ICL) \cite{dong2023survey} in the process. The in-context examples are randomly sampled from $S$. As more prompts are generated and saved to $S$, the pool of available examples also expands, ensuring the diversity of the prompts. We always sample the maximum possible number of in-context examples while satisfying the context length constraint.

\textbf{Prompt Evaluation.}
Similar to generation, we also apply ICL during the evaluation step. We use Llama-2-70B-Chat to evaluate the review-prompt pair based on a 5-point scale with five in-context examples for each score from 1 to 5. The in-context examples (shown in Appendix \ref{app:exp_eval}) are manually constructed and remain consistent across all evaluations. Inspired by chain-of-thought prompting \cite{wei2023chainofthought}, we prompt the LLM to generate an explanation for the score before producing the final score to encourage more accurate assessments.

\textbf{Regeneration.}
To ensure the quality of the generated prompt, the pipeline regenerates the prompt if the score is not $5$. Since the in-context examples for generation are randomly sampled rather than a fixed set, the regeneration step is guaranteed to generate a different prompt compared to the previous generations, minimizing redundancy. We use a limit of $5$ generations per review, and the review is excluded from further generation if it exceeds the limit. $93.60\%$ of the reviews take less than or equal to $3$ generations to reach a score of $5$.

\subsection{Dataset Details}
We incorporate parts of the PeerRead and NLPeer datasets. \textbf{CONLL-16} and \textbf{ACL-17} from PeerRead contain papers and reviews from the NLP domain. 
The reviewing process is double-blind and the formats of the review are unstructured. 
NLPeer's \textbf{COLING-20} and \textbf{ARR-22} are collected via a donation-based workflow in NLP domain with formats in free-form reports and standardized structured review forms.

In addition to the prior datasets, we crawl ICLR papers from 2017 to 2023 through {OpenReview}\footnote{\url{https://openreview.net/}} and NeurIPS papers from 2016 to 2020 through {NeurIPS Proceedings}\footnote{\url{http://papers.neurips.cc/}} and from 2021 to 2022 through OpenReview. The resulting datasets are \textbf{ICLR-17-23} and \textbf{NeurIPS-16-22}. For each paper’s review, we follow the format of the previous datasets to keep as much metadata information as possible including reference and meta reviews from official reviewers, and final decisions.

\textbf{Unification.}
The diverse sources of datasets are converted into a unified format to enhance accessibility and consistency. For each paper, we include the full text of the paper, metadata, and corresponding reviews and prompts. For the contents of the paper, we use {Science Parse}\footnote{\url{https://github.com/allenai/science-parse}} from AllenAI to parse the PDFs of the papers into construct structured JSON files. Each paper is accompanied by detailed metadata, providing essential information about the paper. The detailed sections of paper and metadata are shown in Appendix~\ref{app:data_details}. The reviews contain both textual components and scores that are divided into different sections based on the venue-specific formats. In addition, we employ our \textit{PGE} pipeline to construct a prompt for each review. For simplicity, we only use the text part of the review for prompt generation and review generation.

\textbf{Analysis.}
The statistics of our dataset are shown in Table~\ref{tab:dataset}. Our dataset consists of more than $27k$ papers and $99k$ reviews in various domains. The average paper length spans from $4k$ to $7k$, demonstrating substantial variability. The review length and prompt length exhibit smaller variances, averaging from $400$ to $500$ and $45$ to $60$ respectively.
% vocabulary overlap based on the Jaccard metric on review lemmas of the datasets in NLPEER.
Compared to other review datasets (Table~\ref{tab:dataset_comp}), our dataset has the largest number of papers and reviews and is the only dataset that includes  aspect prompts.

\textbf{Licensing and Personal Data}
All datasets are distributed under an open Creative Commons license and compiled with explicit consent or sourced from materials with an open license. We attribute authors of the papers in our dataset while excluding personal and reviewer metadata.

\begin{table}[t]\centering
\caption{Results of the model variations using three metrics across six venues (SS-E0: \textsc{SingleS-E0}, SS-E: \textsc{SingleS-E}, SS: \textsc{SingleS}, R2-E: \textsc{Reviewer2-E}, R2: \textsc{Reviewer2}). The best-performing model for each venue and metric is highlighted in bold. \label{tab:result}}
\vskip -0.3cm
\resizebox{\linewidth}{!}{
\begin{tabular}{cccccccc} 
\midrule[0.15ex]
& & \multirow{2}{*}{\textbf{Method}} & \multirow{2}{*}{\textbf{BLEU}}  & \multicolumn{3}{c}{\textbf{ROUGE} (max)} & \multirow{2}{*}{\textbf{BertScore}}
\\  
& & & (max) & \textbf{R-1} & \textbf{R-2} & \textbf{R-L} & (max)
\\  \midrule[0.05ex]
\multirow{10}{*}{\rotatebox[origin=c]{90}{In-domain}} & \multirow{5}{*}{\rotatebox[origin=c]{90}{ICLR}} & SS-E0    &  8.15 & 29.93 &  7.14 & 13.76 & 68.45 \\
                      && SS-E   & 12.53 & 39.63 & 10.19 & 19.76 & 79.40 \\
                      && R2-E    & 13.32 & 40.06 & 10.59 & 20.34 & 80.11 \\
                      && SS    & 15.08 & 40.77 & 11.78 & 21.09 & 81.18 \\
                      && R2      & \textbf{16.94} & \textbf{44.58} & \textbf{13.56} & \textbf{22.62} & \textbf{83.61} \\
\cline{2-8}
& \multirow{5}{*}{\rotatebox[origin=c]{90}{NeurIPS}} & SS-E0    &  8.29 & 28.96 &  6.98 & 13.63 & 67.82 \\
                      && SS-E   & 11.72 & 39.54 &  9.75 & 19.67 & 79.17 \\
                      && R2-E    & 12.91 & 39.87 & 10.02 & 19.81 & 80.17 \\
                      && SS    & 14.44 & 40.62 & 11.22 & 20.8  & 81.83 \\
                      && R2      & \textbf{16.24} & \textbf{42.15} & \textbf{13.11} & \textbf{22.52} & \textbf{83.23} \\
\midrule[0.05ex]
\multirow{20}{*}{\rotatebox[origin=c]{90}{Cross-domain}} & \multirow{5}{*}{\rotatebox[origin=c]{90}{ACL}}  & SS-E0    &  5.02 & 30.77 &  6.28 & 12.69 & 68.90 \\
                      && SS-E   &  4.67 & 35.23 &  7.07 & 16.53 & 78.15 \\
                      && R2-E    &  4.82 & 36.44 &  7.98 & 16.73 & 80.03 \\
                      && SS    &  5.40 & 35.73 &  7.94 & 16.94 & 80.25 \\
                      && R2      &  \textbf{6.49} & \textbf{36.88} &  \textbf{8.04} & \textbf{17.77} & \textbf{83.65} \\
\cline{2-8}
&\multirow{5}{*}{\rotatebox[origin=c]{90}{ARR}}  & SS-E0    &  6.01 & 32.48 &  7.89 & 13.91 & 69.34 \\
                      && SS-E  &  6.89 & 38.30 &  9.67 & 18.67 & 79.09 \\
                      && R2-E    &  6.96 & 39.17 & 10.94 & 19.53 & 80.69 \\
                      && SS    &  6.73 & 38.93 & 11.22 & 19.61 & 81.03 \\
                      && R2      &  \textbf{7.46} & \textbf{40.18} & \textbf{12.04} & \textbf{20.76} & \textbf{82.29} \\
\cline{2-8}
&\multirow{5}{*}{\rotatebox[origin=c]{90}{COLING}}& SS-E0    &  3.66 & 30.51 &  6.49 & 12.83 & 69.19 \\
                      && SS-E   &  2.65 & 35.31 &  6.92 & 16.5  & 77.92 \\
                      && R2-E    &  3.01 & 35.09 &  7.34 & 17.74 & 78.15 \\
                      && SS    &  3.34 & 34.57 &  8.11 & 17.14 & 80.21 \\
                      && R2      &  \textbf{4.37} & \textbf{37.13} &  \textbf{9.18} & \textbf{18.91} & \textbf{83.35} \\
\cline{2-8}
&\multirow{5}{*}{\rotatebox[origin=c]{90}{CONLL}}& SS-E0    &  5.18 & 32.01 &  6.32 & 12.75 & 69.45 \\
                      && SS-E   &  3.41 & 35.16 &  6.89 & 16.18 & 78.39 \\
                      && R2-E    &  3.59 & 34.28 &  6.74 & 16.82 & 80.15 \\
                      && SS    &  5.09 & 33.85 &  6.88 & 16.52 & 79.83 \\
                      && R2      &  \textbf{6.07} & \textbf{35.38} &  \textbf{7.40} & \textbf{18.22} & \textbf{83.13} \\
\midrule[0.15ex]
\end{tabular}}
\vskip -0.3cm
\end{table}

\begin{figure}[t]
\centering
\includegraphics[scale=0.50, trim={0 0 0 0},clip]{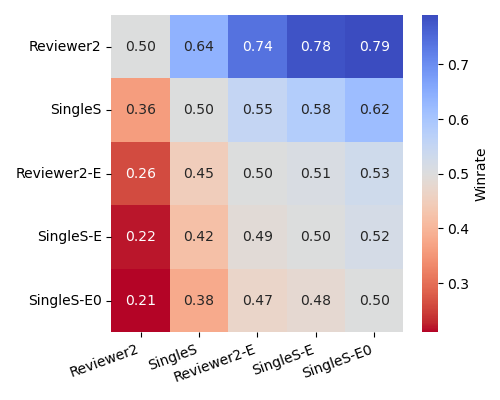}
\vskip -0.3cm
\caption{Pairwise winrates on faithfulness, coverage, coherence, and specificity among \textsc{Reviewer2} and baselines using GPT4 as a judge.}
\label{fig:winrate}
\vskip -0.5cm
\end{figure}

\section{Experiments} \label{variants}

In the following section, we evaluate review quality, review specificity, and aspect coverage as key properties of the generated reviews. We provide extensive ablation experiments that identify how much each novel contribution of our approach contributes to improved performance. In particular, we compare \textsc{Reviewer2} against the following baselines:
\begin{itemize}[leftmargin=*,noitemsep,topsep=0pt]
    \itemsep0em
    \item \textsc{Reviewer2-E}: Following \cite{yuan2021automate}, we apply a cross-entropy (CE) extraction method to extract a diverse set of sentences from the paper to represent the content of the paper. The framework is the same as \textsc{Reviewer2} while we only use the extracted part instead of the full paper:
    $M^E_p: e \rightarrow \{x^1, ..., x^k\}$, $M^E_r: (e, x) \rightarrow y$ where $e$ is the extracted content from paper $p$. This ablation is used to evaluate the difference between using the full paper compared to an extractive summary.
    \item \textsc{SingleS}: We fine-tune a single-stage model to directly generate reviews from the full context of the paper without an aspect prompt, $M^S_r: p \rightarrow y$. Prompts are neither used in fine-tuning nor inference. This ablation is designed to evaluate the effect of aspect prompts.
    \item \textsc{SingleS-E}: This variant involves fine-tuning a single model to generate reviews only from extractive summaries of papers, $M^{SE}_r: e \rightarrow y$. This method aligns with commonly employed pipelines in previous papers and serves as a baseline representing the state-of-the-art.
    \item \textsc{SingleS-E0}: This zero-shot approach prompt an LLM to generate a review from the extracted context directly without aspect prompts. This baseline evaluates the effect of fine-tuning.
\end{itemize}

\begin{figure*}[t]
\centering
\includegraphics[scale=0.50, trim={20 20 15 30},clip]{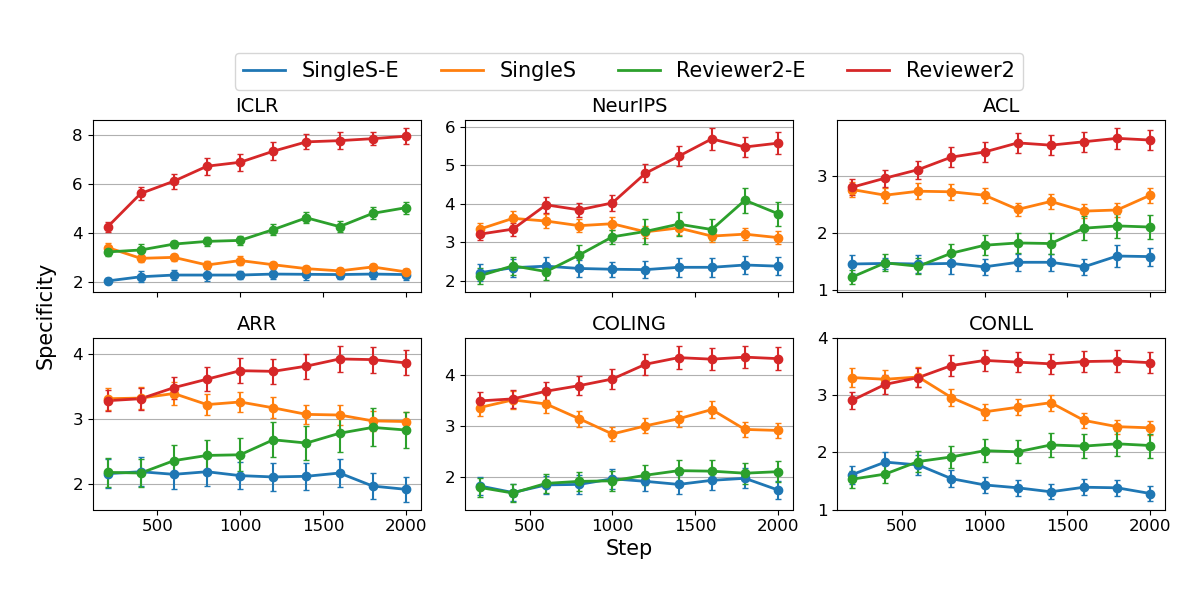}
\vskip -0.3cm
\caption{Specificity plots of four methods for $2000$ steps across six venues. }
\label{fig:specificity}
\vskip -0.3cm
\end{figure*}

%\subsection{Implementation Details}
We use \textsc{SingleS-E} and \textsc{SingleS} as proxies for the baseline methods proposed by \cite{yuan2021automate, Lin_2023}. Llama-2-70B-Chat~\cite{touvron2023llama} is used as the instruction-following model for \textit{PGE} and Llama-2-7B-Chat is used for \textsc{Reviewer2} and the single stage baselines. More experimental details are shown in Appendix~\ref{app:exp_detail}. We randomly select $80\%$ of ICLR and NeurIPS papers for training, $10\%$ for validation, and $10\%$ for testing while using all the papers in other venues for testing. Since the other venues have review formats different from ICLR and NeurIPS, this allows us to test adaptability to different review formats.

\subsection{Quality Analysis}

%\textbf{Metric.} 
To compare the generated reviews with the reference reviews, we employ three metrics: BLEU~\cite{bleu}, ROUGE~\cite{lin-hovy-2003-automatic}, and BertScore~\cite{zhang2020bertscore}. BLEU and ROUGE measure the n-gram similarity while BertScore measures the semantic similarity in the embedding space. Notably, there are several reference reviews for each paper. When computing BLEU, ROUGE, and BertScore, following \cite{yuan2021automate}, we use the maximum value instead of an average since the generated reviews do not need to be closely aligned with all references, given that the reference reviews may focus on different aspects. To compare across the generated reviews, we prompt GPT4 to select the better review based on faithfulness, coverage, coherence, and specificity on $100$ randomly sampled reviews for each method across six venues. More details of the evaluation is shown in Appendix~\ref{app:winrate}.

\textbf{Result.} Table~\ref{tab:result} compares the performance of \textsc{Reviewer2} against several ablations and baselines. Overall, \textsc{Reviewer2} outperforms all methods across all metrics and datasets, demonstrating the effectiveness of leveraging both the full context of the paper and the aspect prompt. The comparisons between \textsc{Reviewer2} and \textsc{SingleS} as well as \textsc{Reviewer2-E} and \textsc{SingleS-E} reveal consistent performance improvement through the two-stage approach. Furthermore, the comparison between \textsc{Reviewer2} and \textsc{Reviewer2-E} shows that avoiding extractive summaries provides an additive benefit on top of using aspect prompts. On the cross-domain datasets (ACL, ARR, COLING, CONLL) we can observe a comparable BertScore with ICLR and NeurlPS using \textsc{Reviewer2}, demonstrating the semantic adaptability of the method to domains that the methods was not trained on. 

Figure~\ref{fig:winrate} compares \textsc{Reviewer2} with the baselines on faithfulness, coverage, coherence, and specificity using GPT4 as a judge. \textsc{Reviewer2} consistently achieves higher winrates compared to the baselines, demonstrating the effectiveness of our method in producing high-quality reviews.

To further illustrate \textsc{Reviewer2}, we included aspect prompts produced by $M_p$ and a review produced by $M_r$ in Appendix~\ref{app:reviewofpaper}. 

\subsection{Specificity Analysis}

%\textbf{Metric.}
A highly specific review identifies specific issues of the given paper, and it does not look like a generic review that could apply to other papers.
To formalize this into a concise metric, we measure the specificity of the review by calculating the drop in BertScore when pairing the review with the reference reviews of a different paper. A generated review with high specificity will lead to a large average drop, while a generic review will lead to a smaller drop. Formally, given papers $P$, reviews $Y$, and generated reviews $\hat{Y}=\{\hat{y}_1, \hat{y}_2, ..., \hat{y}_m\}$, we define specificity (\textsc{Spe $\uparrow$}) as:
%\begin{align*}
%\mathrm{\textsc{Spe}} = \frac{1}{m}\sum_{i=1}^{m}
%                           &\max \{\mathrm{sim}(\hat{y}_i, y_i^n)|1\leq n \leq n_i\} \\
%                           &- \max \{\mathrm{sim}(\hat{y}_j, y_i^n)|1\leq n \leq n_i\} 
%\end{align*}
\begin{align*}
\mathrm{\textsc{Spe}} = \frac{1}{m}&\sum_{i=1}^{m}
                           \max \{\mathrm{sim}(\hat{y}_i, y_i^n)|1\leq n \leq n_i\} \\
                           &\!\!\!\!\!\!\!\!\!\!\!- \frac{1}{m\!-\!1}\sum_{j \not= i}\max \{\mathrm{sim}(\hat{y}_i, y_j^n)|1\leq n \leq n_i\} 
\end{align*}
where $\mathrm{sim}(a, b)$ denotes the BertScore between $a$ and $b$ and $\hat{y}_j$. We approximate the inner sum by Monte Carlo sampling $j \sim [1, m]\setminus i$. %We use the maximum value across references similar to the previous section.

\textbf{Result.} To obtain a reliable measure, we conducted ten random shuffles and calculated the average. The result is shown in Figure~\ref{fig:specificity} along with the variance. For methods that do not make use of aspect prompts, \textsc{SingleS} and \textsc{SingleS-E}, the specificity drops with more training steps. This indicates that increased training without prompts leads to more generic reviews. For the methods that use prompts, \textsc{Reviewer2-E} and \textsc{Reviewer2}, the specificity consistently increases with a higher number of steps. Notably, the difference between \textsc{Reviewer2} and \textsc{SingleS} is higher than the difference between \textsc{Reviewer2-E} and \textsc{SingleS-E}, suggesting that adding prompts on top of the full context leads to higher improvement comparing to adding to the extracted context.

\subsection{Control Analysis}

\begin{table}[t]\centering
\caption{Effect of prompts for \textsc{SingleS} (SS) and \textsc{Reviewer2} (R2) across six venues. \label{tab:effect_of_prompt}}
\vskip -0.3cm
\resizebox{\linewidth}{!}{
\begin{tabular}{ccc} 
\midrule[0.15ex]
\multirow{3}{*}{\rotatebox[origin=c]{90}{avg}} 
& SS  & $\displaystyle\frac{1}{m}\sum_{i=1}^m\frac{1}{n_i}\sum_{n=1}^{n_i}\mathrm{sim}(M_r^S(p_i), y_i^n)$ \\
& R2    & $\displaystyle\frac{1}{m}\sum_{i=1}^m\frac{1}{n^2_i}\sum_{n=1}^{n_i}\sum_{k=1}^{n_i}\mathrm{sim}(M_r(p_i, x_i^n), y_i^k)$\\
\midrule[0.05ex]
\multirow{3}{*}{\rotatebox[origin=c]{90}{max}} 
& SS  &  $\displaystyle\frac{1}{m}\sum_{i=1}^m\max\{\mathrm{sim}(M_r^S(p_i), y_i^n)|$\\[-2ex]
& & \qquad$\displaystyle 1\leq n \leq n_i\}$\\
& R2    &  $\displaystyle\frac{1}{m}\sum_{i=1}^m\frac{1}{n_i}\sum_{n=1}^{n_i}\max\{\mathrm{sim}(M_r(p_i, x_i^n), y_i^k)|$\\[-2ex]
& & \qquad \qquad$\displaystyle 1 \leq k \leq n_i\}$\\
\hline\hline
\end{tabular}}
\resizebox{\linewidth}{!}{
\begin{tabular}{cccccccc}
          & Method & ICLR  & NeurIPS  & ACL   & ARR   & COLING& CONLL \\
\midrule[0.05ex]
\multirow{2}{*}{\rotatebox[origin=c]{90}{avg}} 
& SS  & \textbf{80.19} & 80.23 & \textbf{79.85} & 80.23 & 79.42 & \textbf{78.41} \\
& R2    & 80.13 & \textbf{80.36} & 79.14 & \textbf{79.96} & \textbf{79.53} & 78.28 \\
\midrule[0.05ex]
\multirow{2}{*}{\rotatebox[origin=c]{90}{max}} 
& SS  & 81.18 & 81.83 & 80.25 & 81.03 & 80.21 & 79.83 \\
& R2    & \textbf{83.63} & \textbf{83.41} & \textbf{83.54} & \textbf{82.51} & \textbf{83.19} & \textbf{83.32} \\
\midrule[0.15ex]
\end{tabular}}
\vskip -0.3cm
\end{table}

To assess how responsive \textsc{Reviewer2} is to the aspect prompts, we conduct experiments that compare \textsc{Reviewer2} and \textsc{SingleS}. The $M_r$ model in \textsc{Reviewer2} is given the prompts generated by \textit{PGE}. We compute the average similarity of the generated review to the reference reviews for both methods as well as the maximum similarity. The detailed equations for the computations are shown in Table~\ref{tab:effect_of_prompt}. BertScore is used for computing sim.

\textbf{Result.} \textsc{Reviewer2} and \textsc{SingleS} have similar average similarity while \textsc{Reviewer2} has a higher maximum similarity across all six venues. This means that \textsc{SingleS} generates reviews that are close to all the reference reviews, but that are not particularly close to any one of them. In contrast, \textsc{Reviewer2} is consistently able to generate reviews that closely match one of the references. 
% This provides evidence that \textsc{Reviewer2} is responsive to aspect prompts and can cover the desired aspects.

\begin{table}[t]\centering
\caption{Coverability (\textsc{Cov $\downarrow$}) for \textsc{Reviewer2-E} (R2-E) and \textsc{Reviewer2} (R2) across six venues. \label{tab:control_result}}
\vskip -0.3cm
\resizebox{\linewidth}{!}{
\begin{tabular}{ccccccc} 
\midrule[0.15ex]
Method & ICLR  & NeurIPS  & ACL   & ARR   & COLING& CONLL \\
\midrule[0.05ex]
R2-E  & 13.55 & 12.66 & 16.62 & 15.29 & 14.84 & 15.46 \\
R2    & \textbf{4.22} & \textbf{3.99} & \textbf{3.23} & \textbf{2.91} & \textbf{5.09} & \textbf{4.25} \\
\midrule[0.15ex]
\end{tabular}}
\vskip -0.3cm
\end{table}

\subsection{Coverage Analysis}

Finally, we evaluate whether authors can achieve good coverage through the choice of aspect prompts and the effect of different aspect prompts on generation. Since $M_r$ and $M_r^E$ are the only models that permit aspect prompts, we evaluate the effect of aspect prompts on coverage for these two models. Given papers $P$, reviews $Y$, prompts $X$, we define coverability (\textsc{Cov $\downarrow$}) for $M_r$ as:
\begin{alignat*}{3}
\mathrm{\textsc{Cov}} &= \frac{1}{m}\sum_{i=1}^{m} g_i - h_i \\
h_i &= \frac{1}{n_i(n_i-1)}\sum_{n=1}^{n_i}\sum_{\substack{k=1 \\ k\neq n}}^{n_i} \mathrm{sim}(y_i^n, y_i^k) \\
g_i &= \frac{1}{n_i(n_i-1)}\sum_{n=1}^{n_i}\sum_{\substack{k=1 \\ k\neq n}}^{n_i} 
\begin{aligned}
    \mathrm{sim}(& M_r(p_i, x_i^n), \\
    & M_r(p_i, x_i^k))
\end{aligned}
\end{alignat*}
Here, $h_i$ represents the pairwise similarity among the reference reviews for paper $p_i$ while $g_i$ is the pairwise similarity among generated reviews based on the \textit{PGE} prompts in the dataset. The coverability for $M_r^E$ is defined similarly but with $e_i$ as input instead of $p_i$. We use BertScore to calculate the similarities. A high $g_i$ indicates that the generated reviews are similar despite being generated from different prompts. 

\textbf{Result.} The results are shown in Table~\ref{tab:control_result}. While perfectly reproducing the coverage of the human reviews would imply a value of $0$, $M_r$ exhibits significantly better coverage than $M_r^E$, demonstrating its effectiveness in generating tailored responses across diverse prompts for a given paper and the importance of using full context.

\section{Conclusion}
We propose a two-stage review generation framework that incorporates aspect prompts. Analyses of quality, specificity, and controllability indicate that our method can generate high-quality and specific reviews while being controllable based on the aspect prompt. Furthermore, we develop a new pipeline for annotating review datasets with aspect prompts, and we make this new dataset available.

\section{Limitations}
In this section, we discuss some of the limitations for \textit{PGE} and \textsc{Reviewer2}.

\subsection{Disjoint Processes for Generation}
Our current configuration first uses \textit{PGE} to generate prompts and subsequently fine-tunes \textsc{Reviewer2} with the generated prompts. However, this approach leads to a disjointed process, where prompt generation operates independently of review generation, reducing the effectiveness of the generated prompts. Ideally, the generated prompts should assist alignment during fine-tuning. A possible extension is to integrate the two processes together and refine the generated prompts based on the review generation pipeline.

\subsection{Input Inconsistency}
The input to \textit{PGE} consists of human-written reviews, while \textsc{Reviewer2} also incorporates papers. This distinction arises from the limitation of Llama-2-70B-Chat, which only has a context length of 4,096. Although GPT-4~\cite{openai2023gpt4} supports up to 32,000 context length, the associated cost is high since the average context length of the papers is 6,229. The potential improvement in performance may not be worth the increased cost.

\subsection{Limited Domain Knowledge}
Currently, \textsc{Reviewer2} relies on its pre-trained corpus, assuming that the language model used has adequate domain knowledge. This approach might produce inaccurate reviews for papers that demand substantial in-domain expertise. A potential future work could investigate the effectiveness of second-stage pre-training or domain adaptation using the paper corpus.

% \section{Discussion}
% Other tasks with one-to-many mapping can also use this technique.
% Re pre-training models for different domain
% Datasets from more domains
% Multiple papers as input

\section{Ethics}

Automatic review generation is a complex task and bears a wide range of risks. It is crucial to emphasize that the ongoing efforts in this field are not designed to replace human reviewers; instead, they function as a valuable tool for authors and a guiding resource for human reviewers. This research is an exploratory work within this domain, and it is important to stress that the outcomes produced by the models should not be misconstrued as definitive and authentic reviews of the respective papers. In utilizing datasets, we adhere to the intended purposes outlined in previous works. The datasets we released offer many possibilities for advancing research in NLP, including but not limited to review generation, instruction following, and self-alignment.

\section {Acknowledgment}

This research was supported in part by NSF Awards IIS-2312865 and OAC-2311521. All content represents the opinion of the authors, which is not necessarily shared or endorsed by their respective employers and/or sponsors.

\newpage
\bibliography{custom}

\clearpage
\onecolumn
\appendix
\section{Geometric Intuition of Illustration} \label{app:illustration}.

\begin{figure*}[h]
\centering
\includegraphics[scale=0.7, trim={0 0 0 0},clip]{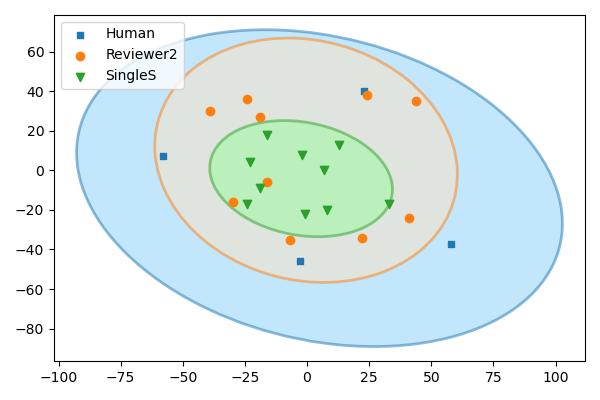}
\vskip -0.3cm
\caption{Embedding visualization of the reviews from human, \textsc{Reviewer2}, and \textsc{SingleS}. The eclipse represents the confidence interval with 2 standard deviations.}
\label{fig:emb_visual}
\vskip -0.3cm
\end{figure*}

To justify the intuition of our illustration in Figure~\ref{fig:prompt_illustration}, we visualize the text embeddings of actual generations from \textsc{Reviewer2} and \textsc{SingleS}. For a given paper, we use \textsc{Reviewer2} and \textsc{SingleS} to generate $10$ reviews respectively. Then, we use SFR-Embedding-Mistral\footnote{Huggingface Model Card: Salesforce/SFR-Embedding-Mistral} as the embedding model to embed the text and visualize the embeddings using t-SNE~\citep{tsne}. The \textsc{SingleS} generations have a much smaller coverage and reside around the middle of the space spanned by the human-written reviews, while generations from \textsc{Reviewer2} are more dispersed and closer to the human references. This plot aligns closely with the intuition we convey in Figure~\ref{fig:prompt_illustration} (b) and (c).

\clearpage
\onecolumn
\section{Prompts for \textit{PGE}} \label{app:prompts}
\begin{table*}[htb]\centering
\vskip -0.3cm
\raggedright{\textbf{Prompt for Generation}}
\resizebox{\linewidth}{!}{
\begin{tabular}{p{0.9\linewidth}} 
\midrule[0.5ex]
   [INST] <<SYS>> You are a helpful, respectful and honest assistant. Always answer as helpfully as possible, while being safe.  Your answers should not include any harmful, unethical, racist, sexist, toxic, dangerous, or illegal content. Please ensure that your responses are socially unbiased and positive in nature. If a question does not make any sense, or is not factually coherent, explain why instead of answering something not correct. If you don't know the answer to a question, please don't share false information. <</SYS>> \newline
   Analyzing the provided review, identify a set of questions that the reviewer is attempting to address regarding the paper without being too specific. \newline
   Here are some examples: \newline \newline
   Review: \newline
   [SAMPLED REVIEW FROM $S$] \newline
   Questions to address: \newline
   [SAMPLED PROMPT FROM $S$] \newline \newline
   Review: \newline
   [SAMPLED REVIEW FROM $S$] \newline
   Questions to address: \newline
   [SAMPLED PROMPT FROM $S$] \newline \newline
   Review: \newline
   [SAMPLED REVIEW FROM $S$] \newline
   Questions to address: \newline
   [SAMPLED PROMPT FROM $S$] \newline \newline
   Review: \newline
   [REVIEW FOR GENERATION] \newline
   Questions to address:[/INST] \\
\midrule[0.5ex]
\end{tabular}}
\end{table*}

\begin{table*}[htb]\centering
\vskip -0.3cm
\raggedright{\textbf{Prompt for Evaluation}}
\resizebox{\linewidth}{!}{
\begin{tabular}{p{1.0\linewidth}} 
\midrule[0.5ex]
   [INST] <<SYS>> You are a helpful, respectful and honest assistant. Always answer as helpfully as possible, while being safe.  Your answers should not include any harmful, unethical, racist, sexist, toxic, dangerous, or illegal content. Please ensure that your responses are socially unbiased and positive in nature. If a question does not make any sense, or is not factually coherent, explain why instead of answering something not correct. If you don't know the answer to a question, please don't share false information. <</SYS>> \newline
   Below is a set of questions and a candidate answer. Evaluate the quality of the questions. Are the questions a good match to the candidate answer?  Please assign a score using the following 5-point scale: \newline
   1: This score indicates that the response deviates significantly from the instruction, providing information or addressing aspects that were not required or specified. \newline
   2: This score suggests that the response is limited in scope, focusing on a small subset of the questions posed in the instruction. It does not comprehensively cover the entire set of questions. \newline
   3: This score indicates that the response covers a substantial portion of the questions outlined in the instruction but falls short of addressing all of them. It suggests a moderate level of completeness. \newline
   4: This score indicates that the response covers most of the questions. However, there is some irrelevant information in the answer that is not asked by any of the questions. \newline
   5: This score indicates that the response is comprehensive, addressing all questions in the instruction without any irrelevant information. \newline \newline
   Here are some examples: \newline \newline
   
   Questions: \newline
   [EXAMPLE PROMPT] \newline
   Answer: \newline
   [EXAMPLE REVIEW] \newline
   Assessment: \newline
   [EXAMPLE ASSESSMENT] \newline
   Score: [EXAMPLE SCORE] \newline \newline
   Questions: \newline
   [EXAMPLE PROMPT] \newline
   Answer: \newline
   [EXAMPLE REVIEW] \newline
   Assessment: \newline
   [EXAMPLE ASSESSMENT] \newline
   Score: [EXAMPLE SCORE] \newline \newline
   Questions: \newline
   [EXAMPLE PROMPT] \newline
   Answer: \newline
   [EXAMPLE REVIEW] \newline
   Assessment: \newline
   [EXAMPLE ASSESSMENT] \newline
   Score: [EXAMPLE SCORE] \newline \newline
   Questions: \newline
   [PROMPT FOR EVALUATION] \newline
   Answer: \newline
   [REVIEW FOR EVALUATION] \newline
   Assessment:[/INST] \\
\midrule[0.5ex]
\end{tabular}}
\end{table*}

\clearpage
\section{Example Review-Prompt Pair} \label{app:exp_prompt}

\begin{table*}[htb]\centering
\vskip -0.3cm
\resizebox{\linewidth}{!}{
\begin{tabular}{p{0.1\linewidth}p{0.9\linewidth}} 
\midrule[0.5ex]
Review & Summary Of The Paper \newline
This paper introduces neural matching fields into semantic correspondence. To the best my knowledge, this approach should be the first method to do the task using implicit neural representation. There are two problems: the computation for 4D matching field and the inference efficiency. Authors provide effect method to address the two problems. \newline
Strengths And Weaknesses \newline
This paper employs implicit neural representation to do semantic correspondence. This should be the major contribution. According to the statement of authors, I can follow the idea easily and this idea should work. The disadvantage of this work is the experiments. There are too many quantitative comparisons. According to the data, the performance of this method seems OK. However, authors should provide more visual experiments to convince readers. \newline
Questions \newline
I only have one concern. Traditional Implicit Neural Representation method such as LIIF and NeRF records images into the weights of neural network. One neural network represents one image or one scene. Does NeMF take a neural network to represent a semantic correspondence or a matching cost. If so, how much time will your method cost to train a network? If not so, what is the difference between your method and other semantic correspondence methods. \newline
Limitations \newline
According to my understand, NeMF takes a network to represent a matching cost. In practice, people need a method to compute different matching cost for different image pairs. How does NeMF to deal with this situation. \\
\midrule[0.3ex]
Questions to address & 
1. What is the focus and contribution of the paper on semantic correspondence? \newline
2. What are the strengths of the proposed approach in terms of neural representation? \newline
3. What are the weaknesses for the experiment section? \newline
4. Do you have any concerns on the semantic correspondence representation? \newline
5. What are the limitations regarding the NeMF approach on matching cost representation? \\
\midrule[0.5ex]
\end{tabular}}
\end{table*}

\clearpage
\section{In-Context Example for Evaluation} \label{app:exp_eval}

\begin{table*}[htb]\centering
\vskip -0.3cm
\resizebox{\linewidth}{!}{
\begin{tabular}{p{0.1\linewidth}p{0.9\linewidth}} 
\midrule[0.5ex]
Questions &
1. What is the main contribution of the paper on dictionary learning? \newline
2. What are the strengths of the paper in the theoretical analysis? \newline
3. Do you have any questions regarding the assumptions, theorems, and algorithm of the paper? \newline
4. Could you access the reproducibility of the paper? \\
\midrule[0.3ex]
Answer &
The paper proposes an alternating minimization algorithm for dictionary learning, and theoretical guarantees are also given. In each step the algorithm first uses an l1, l2 and l\_infty algorithm with thresholding to get an estimate of the coefficients, and then use another gradient step to update the dictionary. \newline
To me two shining points of the paper: \newline
1. Guarantee holds for the overcomplete dictionary. \newline
2. Improved the sparsity level requirement by a factor of log d. \newline
Obviously the NIPS format is too short for the arguments the authors are making, and a lot of details are moved to the appendix. Due to time limit I cannot read all the details of the proof. Below are some questions: \newline
1. In A1 you have a mu-incoherence assumption, but mu is not shown in your theorem 3. Is it hidden somewhere? \newline
2. In assumption B1 you mentioned, and I agree that there is a fast random initialization so that the condition holds. Can you give some details about your initialization procedure and guarantees? \newline
3. How do you handle the permutation invariance of A? \newline
4. In your algorithm 1, line 3, the MUS algorithm has a return, but in your definition (equation 2), the return is not specified. Actually the returned should be theta instead of (theta, t, u). \newline
5. “(w\_k\textasciicircum t is the k\textasciicircum th covariate at step t)”? Why w\_k\textasciicircum t is called the k\textasciicircum th covariate? \newline
6. Any simulation result verifying your convergence rate? \\
\midrule[0.3ex]
Assessment & 
The answer addresses the first question by summarizing the main contribution of the paper. 

For the second question, the answer gives two strong points of the paper in its theoretical justifications. The answer address the third question by providing six different questions convering the assumptions, theorems, and the algorithm of the paper. However, the answer fails to address the fourth question. \\
\midrule[0.3ex]
Score & Since the answer fails to address all of the questions, it receives a score of 3. \\
\midrule[0.5ex]
\end{tabular}}
\end{table*}

\clearpage
\twocolumn
\section{Dataset Details} \label{app:data_details}

\subsection{Paper Contents}
\begin{itemize}[leftmargin=*]
    \itemsep0em
    \item title: title of the paper
    \item authors: list of author names
    \item emails: list of author emails
    \item sections: list of sections of the paper
    \begin{itemize}
        \item heading: heading of the section
        \item text: text of the section
    \end{itemize}
    \item references: list of references of the paper
    \begin{itemize}
        \item title: title of the reference
        \item author: list of author names of the reference
        \item venue: venue of the reference
        \item citeRegEx: citation expression
        \item shortCiteRegEx: short citation expression
        \item year: publication year of the reference
    \end{itemize}
    \item referenceMentions: the location of the reference in the paper
    \begin{itemize}
        \item referenceID: numerical reference id
        \item context: context of the reference in the paper
        \item startOffset: start index of the context
        \item endOffset: end index of the context
    \end{itemize}
    \item year: year of publication
    \item abstractText: abstract of the paper
\end{itemize}

\subsection{Metadata Contents}
\begin{itemize}[leftmargin=*]
    \itemsep0em
    \item id: unique id of the paper
    \item conference: venue for the paper
    \item decision: final decision for the paper (accept/reject)
    \item url: link to the PDF of the paper
    \item review\_url: link to the review of the paper
    \item title: title of the paper
    \item authors: list of the authors of the paper
\end{itemize}

\newpage
\section{Experimental Details} \label{app:exp_detail}

\textsc{Reviewer2} and \textsc{SingleS} have a context length of 32,768 while other models have a 4,096 context length. All of the models excluding \textsc{SingleS-E0} are fine-tuned with $8$ A6000 GPUs using DeepSpeed~\cite{deepspeed} stage 2, batch size $64$, gradient accumulation $8$, and warm-up steps $100$ for $2$ epochs. We use the AdamW optimizer with a learning rate $1e-5$ searched from $[5e-6, 1e-5, 2e-5, 5e-5, 1e-4, 2e-4]$.

We perform supervised fine-tuning (SFT) for \textsc{Reviewer2} and each baseline methods that require training. The setup is detailed below:

\begin{itemize}
   \item \textsc{Reviewer2}
   \begin{itemize}
     \item Prompt generation model ($M_p$)
     \begin{description}
     \item[Model:] Llama-2-7B-Chat
     \item[Input:] Full paper
     \item[Output:] Aspect prompts
     \end{description}
     \item Review generation model ($M_r$)
     \begin{description}
     \item[Model:] Llama-2-7B-Chat
     \item[Input:] Full paper and aspect prompt
     \item[Output:] Review
     \end{description}
   \end{itemize}
   
   \item \textsc{Reviewer2-E}
   \begin{itemize}
     \item Prompt generation model ($M^E_p$)
     \begin{description}
     \item[Model:] Llama-2-7B-Chat
     \item[Input:] Extracted sentences from the paper
     \item[Output:] Aspect prompts
     \end{description}
     \item Review generation model ($M^E_r$)
     \begin{description}
     \item[Model:] Llama-2-7B-Chat
     \item[Input:] Extracted sentences from the paper and aspect prompt
     \item[Output:] Review
     \end{description}
   \end{itemize}
   
   \item \textsc{SingleS}
   \begin{itemize}
     \item Review generation model ($M^S_p$)
     \begin{description}
     \item[Model:] Llama-2-7B-Chat
     \item[Input:] Full paper
     \item[Output:] Review
     \end{description}
   \end{itemize}
   
   \item \textsc{SingleS-E}
   \begin{itemize}
     \item Review generation model ($M^{SE}_p$)
     \begin{description}
     \item[Model:] Llama-2-7B-Chat
     \item[Input:] Extracted sentences from the paper
     \item[Output:] Review
     \end{description}
   \end{itemize}
\end{itemize}

\newpage
\onecolumn
\section{Winrate Details} \label{app:winrate}

We are using \texttt{gpt-4-0613} checkpoint for winrate computations.

\begin{table*}[htb]\centering
\raggedright{\textbf{Prompt for Winrate}}
\resizebox{\linewidth}{!}{
\begin{tabular}{p{1.1\linewidth}} 
\midrule[0.3ex]
Which of the following reviews provide a more accerate review of the given paper? Judge based on faithfulness, coverage, coherence, and specificity.
\newline\newline
\#\#\# Paper: \newline
\{\{Paper\}\}
\newline\newline
\#\#\# Review A: \newline
\{\{reviewa\}\}
\newline\newline
\#\#\# Review B: \newline
\{\{reviewb\}\}
\newline\newline
\#\#\# Instructions: \newline
FIRST provide a comparison of the two reviews, explaining which you prefer and why. SECOND, on a new line, state only "A" or "B" to indicate your choice. Your response should use the format: \newline
Comparison: \textless comparison and explanation \textgreater  \newline
Preferred: \textless ``A'' or ``B''\textgreater \\
\midrule[0.3ex]
\end{tabular}}
\end{table*}

\clearpage
\section{Generated Aspect Prompts and Review for this Paper} \label{app:reviewofpaper}

\begin{table*}[htb]\centering
\vskip -0.3cm
\resizebox{\linewidth}{!}{
\begin{tabular}{p{0.1\linewidth}p{0.9\linewidth}} 
\midrule[0.5ex]
Aspect Prompts &
What is the focus and contribution of the paper on peer review generation?\newline
What are the strengths of the proposed approach, particularly in terms of its two-stage architecture?\newline
What are the weaknesses of the paper regarding its experimental setup and proposed metrics?\newline
Do you have any concerns or suggestions for improving the paper's clarity and organization?\\
\midrule[0.3ex]
Review &
\textbf{Summary Of the Paper.}
This paper proposes a two-stage review generation framework that incorporates aspect prompts. They design a prompt generation with evaluation (PGE) pipeline to generate a variety of high-quality aspect prompts, and they make this new dataset available. They evaluate review quality, specificity, and aspect coverage as key properties of the generated reviews. They provide extensive ablation experiments that identify how far each novel contributions in their approach contributes to improved performance. Finally, they conclude by highlighting the advantages of leveraging both the full context of the paper and the aspect prompt.

\paragraph{Strengths.}
The paper clearly states the problem of existing methods that only generate generic reviews and lacks specificity and control. The authors propose a two-stage architecture that explicitly models the distribution of possible aspects that the review may address. They show that this leads to more detailed reviews that better cover the range of aspects that human reviewers identify in the draft.

\paragraph{Weaknesses.}
The authors evaluate the effectiveness of the proposed method with several baselines. However, the baselines are not very convincing. The paper lacks clear motivation for why they use a two-stage architecture. The paper needs to provide more details on how they evaluate the generated prompts.\\
\midrule[0.5ex]
\end{tabular}}
\end{table*}

% \begin{figure}[htb]
% \centering
% \includegraphics[scale=0.75, trim={330 160 350 110},clip]{./figs/dataset_format.png}
% \vskip -0.3cm
% \caption{Dataset Format}
% \label{fig:dataset_format}
% \vskip -0.5cm
% \end{figure}

% \begin{table}[t]\centering
% \caption{Prompt Metrics. \label{tab:prompt_result}}
% \vskip -0.3cm
% \resizebox{\linewidth}{!}{
% \begin{tabular}{cccccc} 
% \midrule[0.15ex]
% \multirow{2}{*}{\textbf{Venue}} & \multirow{2}{*}{\textbf{BLEU}}  & \multicolumn{3}{c}{\textbf{ROUGE} (max)} & \multirow{2}{*}{\textbf{BS} (max)}
% \\  
%  & (max) & \textbf{R-1} & \textbf{R-2} & \textbf{R-L} & 
% \\  \midrule[0.05ex]
% ICLR  & 59.20 & 69.86 & 44.40 & 54.40 & 92.42 \\
% NeurIPS  & 59.48 & 69.56 & 44.27 & 54.56 & 92.48 \\
% ACL   & 37.43 & 58.16 & 30.58 & 41.54 & 87.38 \\
% ARR   & 49.93 & 68.37 & 41.47 & 52.21 & 92.16 \\
% COLING& 31.60 & 57.81 & 29.53 & 40.67 & 86.93 \\
% CONLL & 37.71 & 58.47 & 30.60 & 41.22 & 88.95 \\
% \midrule[0.15ex]
% \end{tabular}}
% \vskip -0.3cm
% \end{table}

\end{document}